\documentclass{ifacconf}
\usepackage{amsmath}
\usepackage{amssymb}
\usepackage{graphicx}      
\usepackage{natbib}        
\usepackage{enumitem}
\usepackage{verbatim}
\usepackage{tikz}
\usepackage{cite}      

\newtheorem{problem}{Problem}

\newtheorem{lemma}{Lemma}

\newtheorem{remark}{Remark}

\begin{document}
\begin{frontmatter}

\title{Robust Quaternion-based Cooperative  Manipulation without  Force/Torque Information \thanksref{footnoteinfo}} 

\thanks[footnoteinfo]{This work was supported by the H2020 ERC Starting Grand BUCOPHSYS, the Swedish Research Council (VR), the Knut och Alice Wallenberg Foundation and the European Union's Horizon 2020 Research and Innovation Programme under the Grant Agreement No. 644128 (AEROWORKS).}

\author[Auth]{Christos K. Verginis} 
\author[Auth]{Matteo Mastellaro} 
\author[Auth]{Dimos V. Dimarogonas}

\address[Auth]{KTH Royal Institute of Technology, 
   Stockholm, CO 10044 Sweden (e-mail: \{cverginis, matteoma, dimos\}@kth.se)}

\begin{abstract}                
This paper proposes a task-space control protocol for the collaborative
manipulation of a single object by $N$ robotic agents. The proposed methodology is decentralized in the
sense that each agent utilizes information associated with its own
and the object's dynamic/kinematic parameters and no on-line communication
takes place. Moreover, no feedback of the contact forces/torques is required, therefore employment of corresponding
sensors is avoided. An adaptive version of the control scheme is also introduced,
where the agents' and object's dynamic parameters are considered unknown. We also use 
unit quaternions to represent the object's orientation. 
In addition, load sharing coefficients between the agents are employed
and internal force regulation is guaranteed. Finally, experimental studies with two robotic arms verify the validity and effectiveness of the proposed control protocol. 
\end{abstract}


\end{frontmatter}

\section{Introduction} \label{sec:intro}
Multi-agent manipulation has gained a notable amount of attention
lately. Difficult tasks including manipulation of heavy loads that
cannot be handled by a single robotic arm necessitate the employment
of multiple agents. Early works develop control architectures
where the robotic agents communicate
and share information with each other \citep{schneider1992object}, and completely decentralized schemes \citep{liu1996decentralized,liu1998decentralized,zribi1992adaptive,khatib1996decentralized,caccavale2000task} 
where each agent uses only local information or observers \citep{gudino2004control},
avoiding potential communication delays.

Impedance and force/motion control is the most common methodology
utilized in the related literature \citep{schneider1992object,caccavale2008six,heck2013internal,erhart2013adaptive,kume2007coordinated,szewczyk2002planning,tsiamis2015cooperative,ficuciello2014cartesian,ponce2016cooperative,gueaieb2007robust}.
Most of the aforementioned works employ force/torque sensors
to acquire knowledge of the manipulator-object contact forces/torques
which however may result to performance decline due to sensor noise
or mounting difficulties. Force/Torque sensor-free methodologies can be found in \citep{wen1992motion,yoshikawa1993coordinated,liu1996decentralized,kume2007coordinated,zribi1992adaptive}, which have inspired the dynamic modeling in this work.

Another important characteristic is the representation of the agent
and object orientation. The most commonly used tools for orientation representation consist of rotation matrices, Euler
angles and the angle/axis convention. Rotation matrices, however, are rarely used in robotic manipulation tasks due to the difficulty of extracting an error vector from them. Moreover, the mapping from Euler angles and angle/axis values to angular velocities exhibits singularities at certain points, rendering thus these representations incompetent. On the other hand, the representation
using unit quaternions, which is employed in this work, constitutes
a singularity-free orientation representation, without complicating
the control design. Unit quaternions are employed in \citep{campa2006kinematic,caccavale2000task,caccavale2008six,aghili2011self}
for manipulation tasks and in \citep{erhart2016model} for the analysis of the interaction dynamics in cooperative manipulation.

In addition, most of the works in the related literature consider
known dynamic parameters regarding the object and the robotic agents.
However, the accurate knowledge of such parameters, such as masses
or moments of inertia, can be a challenging issue; \citep{liu1998decentralized}
proposes an adaptive control scheme through gain tuning and \citep{caccavale2000task} 
considers the robust pose regulation
problem. The adaptive control of single manipulation tasks with uncertain kinematic and dynamic parameters is tackled in \citep{SLotine_Adaptive_06,TAC_17_adaptive_Wang}.

In \citep{erhart2013adaptive} and \citep{erhart2013impedance} kinematic uncertainties
are considered whereas \citep{erhart2015internal} performs an internal force
and load distribution analysis. In \citep{tsiamis2015cooperative} a leader-follower
scheme is employed, and in \citep{wang2015multi} a decentralized force consensus
algorithm is developed; \citep{murphey2008adaptive} and \citep{chaimowicz2003hybrid}
address the problem employing hybrid control schemes. A formation-control approach is considered in \citep{bai2010cooperative} and a fuzzy control methodology is presented in \citep{Li_fuzzy2015}.
In \citep{petitti2016decentralized} the agent dynamics are not taken into account and \citep{wang2016kinematic} considers a kinematic decentralized approach using force feedback.
 Finally, mobile manipulator approaches are
treated in \citep{sugar2002control,tanner2003nonholonomic,ponce2016cooperative}.

In this paper, we propose a novel nonlinear control scheme for trajectory
tracking of an object rigidly grasped by $N$ robotic agents. The main
novelty of our approach is the \textit{combination} of i) coupled object-agents dynamic formulation
which does not require contact forces/torques measurements from corresponding sensors, ii) an
extension to an adaptive version, where the dynamic parameters of
the object and the agents are considered unknown and iii) the employment
of unit quaternions for the object orientation, avoiding thus potential
representation singularities. Moreover, the overall scheme is decentralized
in the sense that each agent utilizes information regarding only its own state, and internal force regulation can be also guaranteed.
Furthermore, in contrast to the majority of the related literature,
we utilize coefficients for load sharing among the robotic arms, which
may exhibit different power capabilities. To the best of the authors'
knowledge, the integration of the aforementioned
attributes for cooperative manipulation has not been addressed before, and turns out to be a challenging problem, due to the high complexity of the coupled object-agents dynamics.
Finally, experimental studies verify the validity and effectiveness of the proposed framework.

The rest of the paper is organized as follows: Section \ref{sec:Notation-and-Preliminaries}
introduces notation and preliminary background. Section \ref{sec:Problem-Formulation}
describes the problem formulation and the overall system's model.
The control scheme is presented in Section \ref{sec:Main Results} and
Section \ref{sec:Experiments} verifies our approach with an experimental setup. Finally, Section \ref{sec:Conclusion} concludes
the paper.

\section{Notation and Preliminaries}\label{sec:Notation-and-Preliminaries}

\subsection{Notation} \label{subsec:Notation}
The set of positive integers is denoted as $\mathbb{N}$ and, given $n\in\mathbb{N}$, $\mathbb{R}^n$ is the real $n$-coordinate space,
$\mathbb{R}^n_{\geq 0}$ and $\mathbb{R}^n_{> 0}$ are the sets of real $n$-vectors with all elements nonnegative and positive, respectively, and $S^n$ is the $n$-D sphere; $I_n\in\mathbb{R}^{n\times n}_{\geq 0}$ and $0_{n\times m}\in\mathbb{R}^{n\times m}, n,m\in\mathbb{N}$, denote the unit matrix and the matrix with all entries zero, respectively. 
 The vector connecting the origins of coordinate frames $\{A\}$ and $\{B$\} expressed in frame $\{C\}$ coordinates in $3$D space is denoted as $p^{\scriptscriptstyle C}_{{\scriptscriptstyle B/A}}\in{\mathbb{R}}^{3}$. Given $a\in\mathbb{R}^3$, $S(a)\in\mathbb{R}^{3\times3}$ is the skew-symmetric matrix
defined according to $S(a)b = a\times b$. The rotation matrix from $\{A\}$ to $\{B\}$ is denoted as $R_{{\scriptscriptstyle B/A}}\in SO(3)$, where $SO(3)$ is the $3$D rotation group. The angular velocity of frame $\{B\}$ with respect to $\{A\}$, expressed in $\{C\}$, is
denoted as $\omega^{\scriptscriptstyle C}_{\scriptscriptstyle B/A}\in \mathbb{R}^{3}$ and it holds that \citep{siciliano2010robotics} $\dot{R}_{\scriptscriptstyle B/A}=S(\omega^{\scriptscriptstyle A}_{\scriptscriptstyle B/A})R_{\scriptscriptstyle B/A}$. We further denote as $\phi_{\scriptscriptstyle A/B}\in\mathbb{T}^3$ the Euler angles representing the orientation of $\left\{B\right\}$ with respect to $\left\{A \right\}$, where $\mathbb{T}^3$ is the $3$D torus. We also define the set $\mathbb{M} = \mathbb{R}^3\times\mathbb{T}^3$. For notational brevity, when a coordinate frame corresponds to an inertial frame of reference $\left\{I\right\}$, we will omit its explicit notation (e.g., $p_{\scriptscriptstyle B} = p^{\scriptscriptstyle I}_{\scriptscriptstyle B/I}, \omega_{\scriptscriptstyle B} = \omega^{\scriptscriptstyle I}_{\scriptscriptstyle B/I}, R_{\scriptscriptstyle B} = R_{\scriptscriptstyle B/I}$ etc.). Finally, all vector and matrix differentiations will be with respect to an inertial frame $\{I\}$, unless otherwise stated.

\subsection{Unit Quaternions}

Given two frames $\{A\}$ and $\{B\}$, we define a unit quaternion $\xi_{\scriptscriptstyle B/A}=[\eta_{\scriptscriptstyle B/A}, \varepsilon^T_{\scriptscriptstyle B/A}]^T\in S^{3}$ describing the orientation of $\{B\}$ with respect to $\{A\}$, with $\eta_{\scriptscriptstyle B/A}\in\mathbb{R}, \varepsilon_{\scriptscriptstyle B/A}\in S^2$, subject to the constraint $\eta^2_{\scriptscriptstyle B/A} + \varepsilon^T_{\scriptscriptstyle B/A}\varepsilon_{\scriptscriptstyle B/A} = 1$. The relation between $\xi_{\scriptscriptstyle B/A}$ and the corresponding rotation matrix $R_{\scriptscriptstyle B/A}$ as well as the axis/angle representation can be found in \citep{siciliano2010robotics}.
For a given quaternion $\xi_{\scriptscriptstyle B/A}=[
\eta_{\scriptscriptstyle B/A}, \varepsilon^T_{\scriptscriptstyle B/A}]^T\in S^3$, its conjugate, that corresponds to the orientation of $ \{A\}$ with respect to $\{B\}$, is \citep{siciliano2010robotics} $\xi^*_{\scriptscriptstyle B/A}=[\eta_{\scriptscriptstyle B/A}, -\varepsilon^T_{\scriptscriptstyle B/A}]^T \in S^3$.
Moreover, given two quaternions ${\xi_{i}}= [
\eta_i, \varepsilon_i^T]^T,i\in\{ 1,2\}$, the quaternion product is defined as \citep{siciliano2010robotics}
\begin{equation}
\xi_1\otimes\xi_2=\left[\begin{array}{c}
\eta_1\eta_2-\varepsilon_1^T\varepsilon_2 \\
\eta_1\varepsilon_2+\eta_2\varepsilon_1+S(\varepsilon_1)\varepsilon_2
\end{array}\right]\in S^{3}.  \label{eq:quat_prod}
\end{equation}

The time derivative of a quaternion $\xi_{\scriptscriptstyle B/A}=[\eta_{\scriptscriptstyle B/A}, \varepsilon^T_{\scriptscriptstyle B/A}]^T\in S^3$ is given by    \citep{siciliano2010robotics}:
\begin{subequations} \label{eq:propagation rule}
\begin{equation}
\dot{\xi}_{\scriptscriptstyle B/A}=\frac{1}{2}{E(\xi_{\scriptscriptstyle B/A})\omega^{\scriptscriptstyle A}_{\scriptscriptstyle B/A}},\label{eq:propagation rule_1}
\end{equation}
where $E:S^3\to\mathbb{R}^{4\times3}$ is defined as:
\begin{equation}
E(\xi)=\left[\begin{array}{c}
-\varepsilon^T\\
\eta I_3-S(\varepsilon)
\end{array}\right]. \notag
\end{equation}
Finally, it can be shown that $E^T(\xi)E(\xi) = I_3$ and hence 
\begin{equation}
\omega^{\scriptscriptstyle A}_{\scriptscriptstyle B/A} = 2E^T(\xi_{\scriptscriptstyle B/A})\dot{\xi}_{\scriptscriptstyle B/A}. \label{eq:propagation rule_2}
\end{equation}
\end{subequations}

\begin{figure}
\begin{center}
\includegraphics[width = 0.3\textwidth]{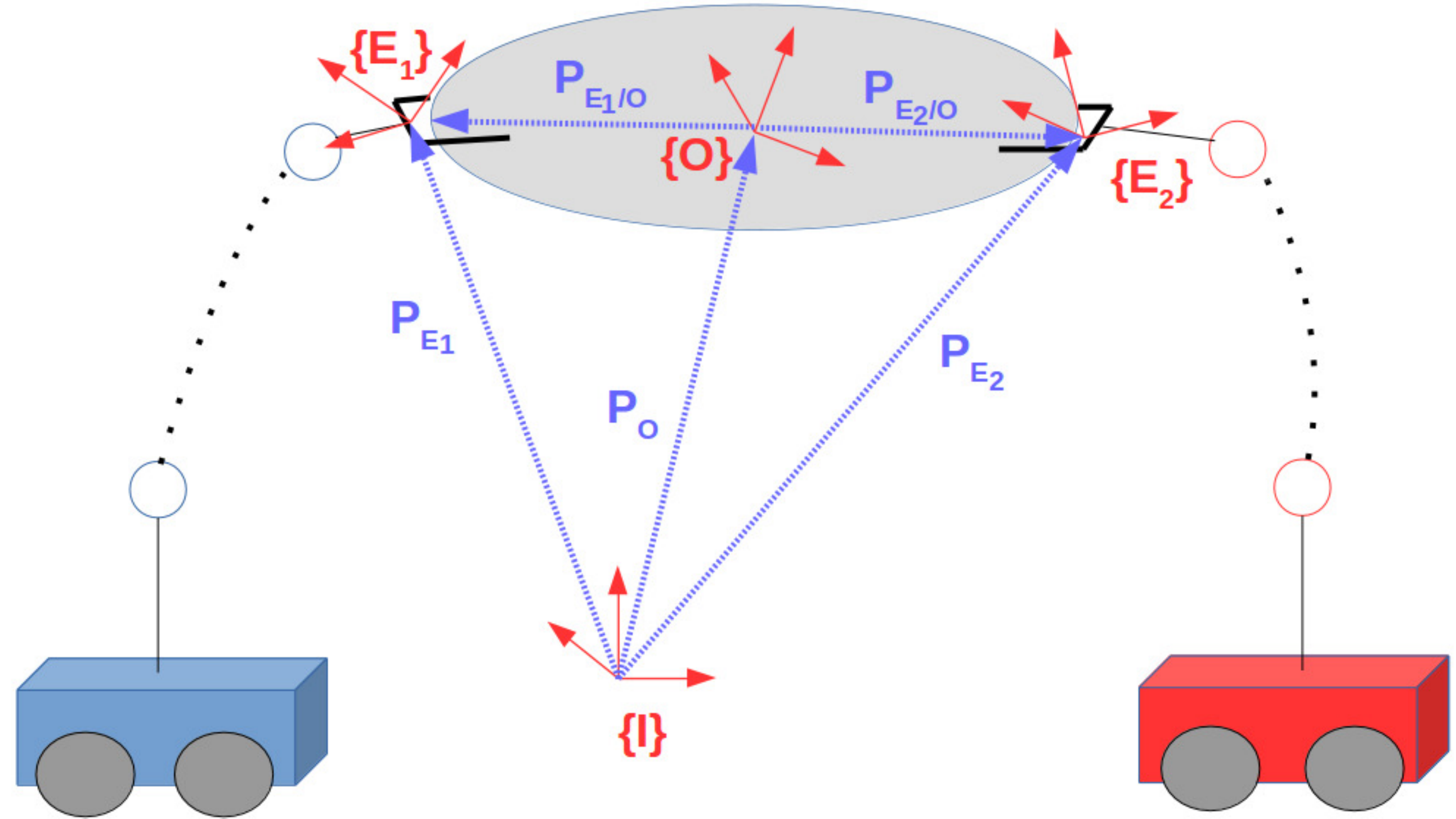}
\caption{Two robotic agents rigidly grasping an object.\label{fig:Two-robotic-arms}}
\end{center}
\end{figure}

\section{Problem Formulation} \label{sec:Problem-Formulation}

Consider $N$ fully actuated robotic agents rigidly grasping an object  
(see Fig. \ref{fig:Two-robotic-arms}). We denote as $q_i\in\mathbb{R}^{n_i}$ the generalized joint-space variables
of the $i$th agent and as $\left\{ E_{i}\right\}$, $\left\{ O\right\}$
the end-effector and
object's center of mass frames, respectively; $\left\{ I\right\} $
corresponds to an inertial frame of reference, as mentioned in Section \ref{subsec:Notation}. The rigidity assumption implies
that the agents can exert both forces and torques along all
directions to the object. We consider that each agent has access to
the position and velocity of its own joint variables and that no interaction force/torque
measurements or on-line information exchange between the agents is
required. Moreover, it is assumed that the desired object profile
as well as relevant geometric features (e.g., center of mass location) are transmitted off-line to
the agents. Finally, we consider that the agents operate
away from kinematic singularity poses \citep{siciliano2010robotics}. In the following, we present the modeling of the coupled kinematics and dynamics of the object and the agents.

\subsection{Kinematics}

In view of Fig. \ref{fig:Two-robotic-arms}, we have that: 
\begin{subequations} \label{eq:coupled_kinematics}
\begin{align}
p_{\scriptscriptstyle E_i}(t) &= p_{\scriptscriptstyle O}(t)+p_{\scriptscriptstyle E_i/O}(q_i) =p_{\scriptscriptstyle O}(t) +  R_{\scriptscriptstyle E_{i}}(q_{i}) p^{\scriptscriptstyle E_i}_{\scriptscriptstyle E_{i}/O},\label{eq:coupled_kinematics_1} \\
\phi_{\scriptscriptstyle E_i}(t) &= \phi_{\scriptscriptstyle O}(t) + \phi_{\scriptscriptstyle E_i/O}, \label{eq:eq:coupled_kinematics_2}
\end{align}
\end{subequations}
$\forall i\in\mathcal{N}$, where $p_{\scriptscriptstyle E_i},\phi_{\scriptscriptstyle E_i},p_{\scriptscriptstyle O},\phi_{\scriptscriptstyle O}$ are the $i$th end-effector's and object's pose, respectively, and $p^{\scriptscriptstyle E_i}_{\scriptscriptstyle E_i/O}$ and $\phi_{\scriptscriptstyle E_i/O}$ are the constant distance and orientation offset between $\{O\}$ and $\{E_i\}$, which are considered known. 
Differentiation of \eqref{eq:coupled_kinematics_1} along with the
fact that, due to the grasping rigidity, it holds that $\omega_{\scriptscriptstyle E_i}=\omega_{\scriptscriptstyle O},$
leads to 
\begin{equation}
v_i(t)=J_{\scriptscriptstyle O_i}(q_i) v_{\scriptscriptstyle O}(t), \label{eq:J_o_i}
\end{equation}
and, by differentiation, to 
\begin{equation}
\dot{v}_i(t)=\dot{J}_{\scriptscriptstyle O_i}(q_i,\dot{q}_i) v_{\scriptscriptstyle O}(t) + J_{\scriptscriptstyle O_i}(q_i) \dot{v}_{\scriptscriptstyle O}(t), \label{eq:J_o_i_dot}
\end{equation}
where $v_{\scriptscriptstyle O}, v_i:\mathbb{R}_{\geq 0}\to\mathbb{R}^6$ with $v_{\scriptscriptstyle O}(t) = [  
\dot{p}^T_{\scriptscriptstyle O}(t), \omega^T_{\scriptscriptstyle O}(t)]^T$,$v_i(t)$ $= [\dot{p}^T_{\scriptscriptstyle E_{i}}(t), \omega^T_{\scriptscriptstyle E_{i}}(t)]^T$ are the object's center of mass' and end-effectors'
velocities respectively. Also, $J_{\scriptscriptstyle O_i}:\mathbb{R}^{n_i}\to\mathbb{R}^{6\times6}$ is the object-to-agent Jacobian matrix, with
\begin{equation}
J_{\scriptscriptstyle O_i}(q_i)=\left[\begin{array}{cc}
I_{ 3} & S(p_{\scriptscriptstyle O/E_{i}}(q_i))\\
0_{3\times 3} & I_{3}
\end{array}\right],	\label{eq:J_o_i_def}
\end{equation}
which is always full-rank due to the grasp rigidity. 

\begin{remark}
Each agent $i$ can compute $p_{\scriptscriptstyle E_i}, \phi_{\scriptscriptstyle E_i}$ and $v_i$ via its forward and differential kinematics \citep{siciliano2010robotics} $p_{\scriptscriptstyle E_i}(t) = k_{p_i}(q_i),\phi_{\scriptscriptstyle E_i}(t) = k_{\eta_i}(q_i)$ and $v_i(t) = J_i(q_i)\dot{q}_i$, respectively, where $k_{p_i}:\mathbb{R}^{n_i}\to\mathbb{R}^3 ,k_{\eta_i}:\mathbb{R}^{n_i}\to\mathbb{T}^3$ are the forward kinematics and $J_i:\mathbb{R}^{n_i}\to\mathbb{R}^{6\times6}$ is the geometric Jacobian of agent $i\in\mathcal{N}$. In addition, since $p^{\scriptscriptstyle E_i}_{\scriptscriptstyle E_i/O}$ and $\phi_{\scriptscriptstyle E_i/O}$ are known, $p_{\scriptscriptstyle O}, \phi_{\scriptscriptstyle O}$ and $v_{\scriptscriptstyle O}$ can be computed by inverting \eqref{eq:coupled_kinematics} and \eqref{eq:J_o_i}, respectively, without employing any sensory data for the object's configuration. Moreover, from $\phi_{\scriptscriptstyle O}$, we can compute the unit quaternion $\xi_{\scriptscriptstyle O}$ \citep{siciliano2010robotics} to represent the object's orientation, since the desired pose for the object's center of mass will be given in terms of a desired position trajectory $p_{\scriptscriptstyle O,\text{d}}(t)$ and a desired quaternion trajectory $\xi_{\scriptscriptstyle O,\text{d}}(t)$.	
\end{remark}

\subsection{Dynamics }

Next, we consider the following second order dynamics for the object, which can be derived based on
the Newton-Euler formulation: 
\begin{equation}
M_{\scriptscriptstyle O}(x_{\scriptscriptstyle O})\dot{v}_{\scriptscriptstyle O}+C_{\scriptscriptstyle O}(x_{\scriptscriptstyle O},\dot{x}_{\scriptscriptstyle O})v_{\scriptscriptstyle O}+g_{\scriptscriptstyle O}(x_{\scriptscriptstyle O}) = f_{\scriptscriptstyle O}, \label{eq:object dynamics}
\end{equation}
where $x_{\scriptscriptstyle O}:\mathbb{R}_{\geq 0}\to\mathbb{M}$, with $x_{\scriptscriptstyle O}(t) = [p^T_{\scriptscriptstyle O}(t), \phi^T_{\scriptscriptstyle O}(t)]^T$, $M_{\scriptscriptstyle O}:\mathbb{M}\to\mathbb{R}^{6\times6}$ is the positive definite inertia matrix, $C_{\scriptscriptstyle O}:\mathbb{M}\times\mathbb{R}^{6}\to\mathbb{R}^{6\times6}$ is the Coriolis matrix, $g_{\scriptscriptstyle O}:\mathbb{M}\to\mathbb{R}^6$ is the gravity vector, and $f_{\scriptscriptstyle O}\in\mathbb{R}^6$ is the vector of generalized forces acting on the object's center of mass. 

The task space agent dynamics are given by \citep{siciliano2010robotics}:
\begin{equation}
M_i(q_i)\dot{v}_i+C_i(q_i,\dot{q}_i)v_i+ g_i(q_i) =u_i-f_i,\label{eq:manipulator dynamics}
\end{equation}
where $M_i:\mathbb{R}^{n_i}\to\mathbb{R}^{6\times6}$ is the positive definite inertia matrix, $C_i:\mathbb{R}^{n_i}\times\mathbb{R}^{n_i}\to\mathbb{R}^{6\times6}$ is the Coriolis matrix, $g_i:\mathbb{R}^{n_i}\to\mathbb{R}^{6}$ is the task-space gravity term, $f_i\in\mathbb{R}^{6}$ is the vector of generalized forces that agent $i$ exerts on the grasping point with the object and $u_i$ is the task space wrench acting as the control input, $\forall i\in\mathcal{N}$.

The agent dynamics \eqref{eq:manipulator dynamics} can be written in vector form as:
\begin{equation}
{M}({q})\dot{{v}}+{C}({q},\dot{{q}}){v}+ {g}({q}) = {u}-{f},\label{eq:manipulator dynamics_vector_form}
\end{equation}
where ${q} = [[q^T_i]_{i\in\mathcal{N}}]^T\in\mathbb{R}^n$, with $n = \sum_{i\in\mathcal{N}}n_i$, ${v} = [[v^T_i]_{i\in\mathcal{N}}]\in\mathbb{R}^{6N} ,{M} = \text{diag}\{[M_i]_{i\in\mathcal{N}}\}\in\mathbb{R}^{6N\times6N}, {C} = \text{diag}\{[C_i]_{i\in\mathcal{N}}\}\in\mathbb{R}^{6N\times6N} , {f} = [[f^T_i]_{i\in\mathcal{N}}]^T\in\mathbb{R}^{6N}, {u} = [[u^T_i]_{i\in\mathcal{N}}]^T\in\mathbb{R}^{6N}, {g} = [[g^T_i]_{i\in\mathcal{N}}]^T\in\mathbb{R}^{6N}$.

\begin{remark} 
The task space wrench $u_i$ can be translated to joint space inputs $\tau_i\in\mathbb{R}^{n_i}$ via $\tau_i = J^T_i(q_i)u_i + (I_{n_i} - J^T_i(q_i)\bar{J}^T_i(q_i))\tau_{i0}$, where $\bar{J}_i$ is a generalized inverse of $J_i$ \citep{siciliano2010robotics};
$\tau_{i0}$ concerns redundant agents ($n_i > 6$) and does not contribute to end-effector forces.
\end{remark}

Moreover, the following property holds: 
\begin{lemma}
\citep{siciliano2010robotics} The matrices $\dot{M}_{\scriptscriptstyle O}-2C_{\scriptscriptstyle O}$ and $\dot{M}_i-2C_i$
are skew-symmetric.\label{lem:skew-symmetry object+manipulator} 
\end{lemma}

The kineto-statics duality \citep{siciliano2010robotics} along with the grasp rigidity suggest that the force $f_{\scriptscriptstyle O}$ acting on the object's center of mass and the generalized forces $f_i,i\in\mathcal{N}$, exerted by the agents at the grasping points, are related through:
\begin{equation}
f_{\scriptscriptstyle O}=G^T({q}){f},\label{eq:grasp matrix}
\end{equation}
where $G:\mathbb{R}^n\to\mathbb{R}^{6N\times6}$ is the full column-rank grasp matrix, with 
\begin{equation}
G({q}) = [J^T_{\scriptscriptstyle O_1}(q_1),\dots,J^T_{\scriptscriptstyle O_N}(q_N)]^T. \notag
\end{equation}

By substituting \eqref{eq:manipulator dynamics_vector_form} into \eqref{eq:grasp matrix}, we obtain:
\begin{equation}
f_{\scriptscriptstyle O} = G^T({q})\left({u} - {M}({q})\dot{{v}} - {C}({q},\dot{{q}}){v} - {g}({q})  \right),
\end{equation}
which, after substituting \eqref{eq:J_o_i}, \eqref{eq:J_o_i_dot}, \eqref{eq:object dynamics}, and rearranging terms,
yields the overall system coupled dynamics: 
\begin{equation}
\tilde{M}({q},x_{\scriptscriptstyle O})\dot{v}_{\scriptscriptstyle O}+\tilde{C}({q},\dot{{q}},x_{\scriptscriptstyle O},\dot{x}_{\scriptscriptstyle O})v_{\scriptscriptstyle O}+\tilde{g}({q},x_{\scriptscriptstyle O})  = G^T({q}){u},\label{eq:coupled dynamics}
\end{equation}
where 
\begin{subequations} \label{eq:coupled terms}
\begin{align}
\tilde{M}  = & M_{\scriptscriptstyle O}+G^T{M}G \label{eq:coupled M}\\
\tilde{C}  = & C_{\scriptscriptstyle O}+G^T{C}G + G^T{M}\dot{G} \label{eq:coupled C}\\
\tilde{g}  = & g_{\scriptscriptstyle O}+G^T{g}.\label{eq:coupled g} 
\end{align}
\end{subequations}
Moreover, the following Lemma holds. 
\begin{lemma} \label{lem:coupled dynamics skew symmetry}
The matrix $\tilde{M}$ is symmetric and positive
definite and the matrix $\dot{\tilde{M}}-2\tilde{C}$
is skew-symmetric.
\end{lemma}

\begin{pf}
By employing the definition of ${M}$ and the positive definiteness of $M_i,\forall i\in\mathcal{N}$,
it is straightforward to prove the positive definiteness
of ${M}$. Then, in view of \eqref{eq:coupled M}
and by invoking the positive definiteness of $M_{\scriptscriptstyle O}$ and the fact that $G$ is full column-rank,
we deduce the positive definiteness of ${\tilde{M}}.$

Regarding the skew symmetry of $\dot{\tilde{M}}-2\tilde{C},$
notice first that the definitions of ${M}, {C}$ as well as
Lemma \ref{lem:skew-symmetry object+manipulator} imply the skew-symmetry
of $\dot{{M}}-2{C}$.
Moreover, by defining $A=\dot{G}^T{M}\dot{G}$,
we have from \eqref{eq:coupled M}, \eqref{eq:coupled C}:
\begin{equation}
\dot{\tilde{M}}-2\tilde{C}=\dot{M}_{\scriptscriptstyle O}-2C_{\scriptscriptstyle O}+G^T(\dot{{M}}-2{C})G+A-A^T, \notag
\end{equation}
from which, by employing Lemma \ref{lem:skew-symmetry object+manipulator},
we obtain: $(\dot{\tilde{M}}-2\tilde{C})^T = -(\dot{\tilde{M}}+2\tilde{C})$,
which completes the proof. 
\end{pf}
Formally, the problem treated in this paper is the following: 
\begin{problem}
Given a desired bounded object pose specified by $p_{\scriptscriptstyle O,\text{d}}(t)\in\mathbb{R}^3,
\xi_{\scriptscriptstyle O,\text{d}}(t) = [\eta_{\scriptscriptstyle O,\text{d}},\varepsilon^T_{\scriptscriptstyle O,\text{d}}]^T\in S^3$, with bounded first
and second derivatives, find ${u}$ in \eqref{eq:coupled dynamics}
that achieves $\lim\limits_{t\rightarrow\infty}\left[\begin{array}{c}
p_{\scriptscriptstyle O}(t)\\
\xi_{\scriptscriptstyle O}(t)
\end{array}\right]=\left[\begin{array}{c}
p_{\scriptscriptstyle O,\text{d}}(t)\\
\xi_{\scriptscriptstyle O,\text{d}}(t)
\end{array}\right]$.\label{prob:problem1} 
\end{problem}

\section{Main Results}\label{sec:Main Results}
We need first to define the errors associated with the object pose and the desired pose trajectory. We first define the position error $e_p:\mathbb{R}_{\geq 0}\to\mathbb{R}^3$:
\begin{equation}
e_p(t)  = p_{\scriptscriptstyle O}(t)-p_{\scriptscriptstyle O,\text{d}}(t). \label{eq:position error}
\end{equation}
Since unit quaternions do not form a vector space, they cannot be subtracted to form an orientation error; instead we should use the properties of the quaternion group algebra. Let $e_\xi = [e_{\eta}, e^T_{\varepsilon}]^T:\mathbb{R}_{\geq 0}\to S^3$ be the unit quaternion describing the orientation error. Then, it holds that \citep{siciliano2010robotics},
\begin{equation}
e_{\xi}(t) = \xi_{\scriptscriptstyle O,\text{d}}(t)\otimes\xi_{\scriptscriptstyle O}^{*}(t) = 
\begin{bmatrix}
\eta_{\scriptscriptstyle O,\text{d}}(t) \\ \varepsilon_{\scriptscriptstyle O,\text{d}}(t)
\end{bmatrix} \otimes
\begin{bmatrix}
\eta_{\scriptscriptstyle O}(t) \\ -\varepsilon_{\scriptscriptstyle O}(t)
\end{bmatrix}, \notag
\end{equation}
which, by using \eqref{eq:quat_prod}, becomes:
\begin{align}
e_\xi(t) &= \begin{bmatrix}
e_\eta(t) \\ e_\varepsilon(t) 
\end{bmatrix} = \notag \\ &=
\begin{bmatrix}
\eta_{\scriptscriptstyle O}(t)\eta_{\scriptscriptstyle O,\text{d}}(t)+\varepsilon^T_{\scriptscriptstyle O}(t)\varepsilon_{\scriptscriptstyle O,\text{d}}(t) \\
\eta_{\scriptscriptstyle O}(t)\varepsilon_{\scriptscriptstyle O,\text{d}}(t)-\eta_{\scriptscriptstyle O,\text{d}}(t)\varepsilon_{\scriptscriptstyle O}(t)+S(\varepsilon_{\scriptscriptstyle O})(t)\varepsilon_{\scriptscriptstyle O,\text{d}}(t) \label{eq:quat_error}
\end{bmatrix}.
\end{align}
By taking the time derivative of \eqref{eq:position error} and \eqref{eq:quat_error}, employing \eqref{eq:propagation rule} and certain properties of skew-symmetric matrices \citep{campa2006kinematic}, it can be shown that \citep{siciliano2010robotics}
\begin{subequations} \label{eq:error_dynamics}
\begin{align}
\dot{e}_{p}(t) & =  \dot{p}_{\scriptscriptstyle O}(t)-\dot{p}_{\scriptscriptstyle O,\text{d}}(t) \label{eq:position error dynamics}\\
\dot{e}_{\eta}(t) & = \tfrac{1}{2}e^T_{\varepsilon}(t)e_\omega(t) \label{eq:eta_error dynamics} \\
\dot{e}_{\varepsilon}(t) & =  -\tfrac{1}{2}\left(e_\eta(t) I_3 + S(e_{\varepsilon}(t)) \right)e_\omega(t)- S(e_{\varepsilon}(t))\omega_{\scriptscriptstyle O,\text{d}}(t),\label{eq:epsilon error dynamics}
\end{align}
\end{subequations}
where $e_\omega:\mathbb{R}_{\geq 0}\to\mathbb{R}^3$, with $e_\omega(t) = \omega_{\scriptscriptstyle O}(t) - \omega_{\scriptscriptstyle O,\text{d}}(t)$ and $\omega_{\scriptscriptstyle O,\text{d}}(t) = 2E^T(\xi_{\scriptscriptstyle O,\text{d}}) \dot{\xi}_{\scriptscriptstyle O,\text{d}}(t)$, as indicated by \eqref{eq:propagation rule_2}. 

Notice that, considering the properties of unit quaternions, when $\xi_{\scriptscriptstyle O} = \xi_{\scriptscriptstyle O,\text{d}}$, then $e_{\xi}(t) = [1, 0_{1\times3}]^T\in S^3$. If $\xi_{\scriptscriptstyle O} = -\xi_{\scriptscriptstyle O,\text{d}}$, then $e_{\xi}(t) = [-1, 0_{1\times3}]^T\in S^3$, which, however, represents the same orientation. Therefore,  the control objective established in Problem \ref{prob:problem1} is equivalent to 
\begin{equation}
\lim\limits_{t\to\infty} 
\begin{bmatrix}
e_p(t) \\ \lvert e_\eta(t) \rvert \\ e_\varepsilon(t)
\end{bmatrix} = 
\begin{bmatrix}
0_{3\times 1} \\ 1 \\ 0_{3\times 1}
\end{bmatrix}. \label{eq:errors to zero}
\end{equation}

Next, we design control protocols such that the specification \eqref{eq:errors to zero} is met. Firstly, we consider that the dynamics parameters of the object and the agents are known. Then, we extend the proposed scheme to also compensate for unknown dynamic parameters, using adaptive control techniques \citep{Slotine_adaptive87,siciliano2010robotics}. 

\subsection{Non-Adaptive Control Scheme} 

Define the velocity reference signals $v^{\scriptscriptstyle r}_{\scriptscriptstyle O}:\mathbb{R}_{\geq 0}\to\mathbb{R}^6$: 
\begin{align}
v_{\scriptscriptstyle O}^{\scriptscriptstyle r}(t) & =  \begin{bmatrix}
\dot{p}_{\scriptscriptstyle O}^{\scriptscriptstyle r}(t)\\
\omega_{\scriptscriptstyle O}^{\scriptscriptstyle r}(t)
\end{bmatrix} \notag \\
 & =  \begin{bmatrix}
\dot{p}_{\scriptscriptstyle O,\text{d}}(t)-k_p e_p(t)\\
\omega_{\scriptscriptstyle O,\text{d}}(t)-k_{\varepsilon}e_\eta(t)e_\varepsilon(t)
\end{bmatrix} = v_{\scriptscriptstyle O,\text{d}}(t) - Ke(t),\label{eq:reference signals}
\end{align}
where $v_{\scriptscriptstyle O,\text{d}}(t) = [\dot{p}^T_{\scriptscriptstyle O,\text{d}}(t), \omega^T_{\scriptscriptstyle O,\text{d}}(t)]^T\in\mathbb{R}^6$, $k_p,k_\varepsilon\in\mathbb{R}_{>0}$, $K=\text{diag}\{ k_pI_3,k_\varepsilon I_3 \} \in\mathbb{R}_{\geq 0}^{6\times6}$ 
and $e(t)= [e_p^T(t), -e_\eta(t) e^T_{\varepsilon}(t)]^T\in\mathbb{R}^6$.

Furthermore, define the reference velocity error $e_v:\mathbb{R}_{\geq 0}\to\mathbb{R}^6$ as:
\begin{equation}
e_v(t)=v_{\scriptscriptstyle O}(t)-v_{\scriptscriptstyle O}^{\scriptscriptstyle r}(t),\label{eq:reference error}
\end{equation}
and design the decentralized control law for $u_i:\mathbb{R}_{\geq 0}\to\mathbb{R}^{6}$ in \eqref{eq:coupled dynamics}, $i\in\mathcal{N}$, as: 
\begin{equation}
u_i(t) = \mu_i(t) + f_{i,\text{d}}(t)  \label{eq:control law}
\end{equation}
where 
\begin{align}
 \mu_i(t)   = & g_i+\left(C_i J_{\scriptscriptstyle O_i} + M_i\dot{J}_{\scriptscriptstyle O_i}\right)v_{\scriptscriptstyle O}^{\scriptscriptstyle r}(t) + M_iJ_{\scriptscriptstyle O_i}\dot{v}_{\scriptscriptstyle O}^{\scriptscriptstyle r}(t) \notag \\
&  -J^{-T}_{\scriptscriptstyle O_i}\left(k_{v_i}e_v(t)+c_ie(t)\right) \notag \\
f_{i,\text{d}}(t)  =&  c_i J_{\scriptscriptstyle O_i}^{-T}\left(M_{\scriptscriptstyle O}\dot{v}_{\scriptscriptstyle O}^{\scriptscriptstyle r}(t)+C_{\scriptscriptstyle O}v_{\scriptscriptstyle O}^{\scriptscriptstyle r}(t)+g_{\scriptscriptstyle O}\right), \notag
\end{align}
$k_{v_i}\in\mathbb{R}_{>0}$ is a positive gain, $c_i\in\mathbb{R}_{\geq 0}$
are load sharing coefficients with $0\leq c_{i}\leq1,\forall i\in\mathcal{N}, \sum_{i\in\mathcal{N}}c_i=1$, and we have also exploited the dependence of $q_i, \dot{q}_i, x_{\scriptscriptstyle O}, \dot{x}_{\scriptscriptstyle O}$ on time.

The control law (\ref{eq:control law}) can be also written in vector
form: 
\begin{equation}
{u}={\mu}+{f}_{\text{d}},\label{eq:control law vector form}
\end{equation}
where 
\begin{align}
{\mu} = & {g}+\left({C}G +{M}\dot{G}\right)v_{\scriptscriptstyle O}^{\scriptscriptstyle r}(t)+ {M}G\dot{v}_{\scriptscriptstyle O}^{\scriptscriptstyle r}(t)\notag\\
& -\tilde{G}^T\left({K}_v e_v(t)+C_{f}e(t) \right) \notag\\
{f}_{\text{d}} = & \tilde{G}^T C_f\left(M_{\scriptscriptstyle O}\dot{v}_{\scriptscriptstyle O}^{\scriptscriptstyle r}(t)+C_{\scriptscriptstyle O}v_{\scriptscriptstyle O}^{\scriptscriptstyle r}(t)+g_{\scriptscriptstyle O}\right), \notag
\end{align}
${K}_v=[k_{v_1}I_6,\dots,k_{v_N}I_6]^T\in\mathbb{R}^{6N\times6}$, $C_{f}=[c_1 I_6,\dots, c_NI_6]^T$ $\in\mathbb{R}^{6N\times6}$, $f_{\text{d}}= [[f^T_{i,\text{d}}]_{i\in\mathcal{N}}]^T\in\mathbb{R}^{6N}$, and finally, $\tilde{G} = \text{diag}\{[J^{-1}_{\scriptscriptstyle O_i}]_{i\in\mathcal{N}}\}\in\mathbb{R}^{6N\times 6N}$.

By employing the fact that $G^T \tilde{G}^T = [I_6,\dots,I_6] \in\mathbb{R}^{6\times6N}$ as well as $\sum_{i\in\mathcal{N}}c_i = 1$, we multiply \eqref{eq:control law vector form} by $G^T$ to obtain: 
\begin{equation}
G^T {u} =  \tilde{M}\dot{v}^{\scriptscriptstyle r}_{\scriptscriptstyle O}(t) + \tilde{C}v^{\scriptscriptstyle r}_{\scriptscriptstyle O}(t) + \tilde{g} - \sum_{i\in\mathcal{N}}k_{v_i} e_v(t) - e(t), \label{eq:control_3}
\end{equation}
that will be used in the sequel.

The following theorem summarizes the main results of this subsection. 
\begin{thm}
Consider $N$ robotic agents rigidly grasping an object with coupled
dynamics described by (\ref{eq:coupled dynamics}) under the control
protocol (\ref{eq:control law vector form}). Then, under the assumption $e_\eta(0) \neq 0$, the object pose
converges asymptotically to the desired one with all closed loop signals
being bounded, i.e., Problem \ref{prob:problem1} is solved. 
\end{thm}

\begin{pf}
Consider the positive definite and radially unbounded
Lyapunov function : 
\begin{align}
V(e_p,e_\eta,e_{v},t)=&\frac{1}{2}{e_{p}^{T}}{e_{p}}+e_\eta^{2}-1+{e_{\varepsilon}^{T}}{e_{\varepsilon}}\notag\\ &+ \frac{1}{2}e_v^T\tilde{M}(q(t),x_{\scriptscriptstyle O}(t))e_v.\notag
\end{align}
By differentiating $V$ with respect to time and substituting the
error dynamics \eqref{eq:error_dynamics}, we obtain: 
\begin{align}
\dot{V} = & e_p^T\dot{e}_p+e_\eta e_\varepsilon^Te_\omega +e_v^T \tilde{M} \dot{e}_v+\tfrac{1}{2}e_v^T \dot{\tilde{M}}e_v \notag \\
 = & e_p^T(\dot{p}_{\scriptscriptstyle O}-\dot{p}_{\scriptscriptstyle O,\text{d}})+e_\eta e_\varepsilon^T (\omega_{\scriptscriptstyle O}-\omega_{\scriptscriptstyle O,\text{d}})+ \notag\\
 & e_v^T \tilde{M}(\dot{v}_{\scriptscriptstyle O}-\dot{v}_{\scriptscriptstyle O}^{\scriptscriptstyle r})+\frac{1}{2}e_v^T \dot{\tilde{M}}e_v, \notag
\end{align}
from which, in view of \eqref{eq:reference signals},
\eqref{eq:reference error} and \eqref{eq:coupled dynamics},
we derive: 
\begin{align}
\dot{V}  =&  -{e^{T}Ke}+{e^{T}e_{v}}+{e_{v}^{T}}\left(G^T{u}-{\tilde{C}v_{{\scriptscriptstyle O}}-\tilde{g}}\right)- \notag\\
 & {e_{v}^{T}}{\tilde{M}}{\dot{v}_{{\scriptscriptstyle O}}^{{\scriptscriptstyle r}}}+\frac{1}{2}{e_{v}^{T}}{\dot{\tilde{M}}}{e_{v}} \notag \\
 =& -{e^{T}Ke}+{e^{T}e_{v}}+{e_{v}^{T}}\left(G^T{u}-{\tilde{C}v_{{\scriptscriptstyle O}}^{{\scriptscriptstyle r}}-\tilde{g}}\right)- \notag \\
 & {e_{v}^{T}}{\tilde{M}}{\dot{v}_{{\scriptscriptstyle O}}^{{\scriptscriptstyle r}}}+{e_{v}^{T}}\left(\frac{1}{2}{\dot{\tilde{M}}}-{\tilde{C}}\right){e_{v}}. \notag
\end{align}
Then, by employing Lemma \ref{lem:coupled dynamics skew symmetry}, $\dot{V}$
becomes: 
\begin{equation}
\dot{V}  = -{e^{T}Ke}-{e_{v}^{T}}(-{e}+{\tilde{M}}{\dot{v}_{{\scriptscriptstyle O}}^{{\scriptscriptstyle r}}}+{\tilde{C}v_{{\scriptscriptstyle o}}^{{\scriptscriptstyle r}}+\tilde{g}-}G^T{u}), \notag
\end{equation}
and after substituting \eqref{eq:control_3}:
\begin{equation}
\dot{V}= -k_pe_p^Te_p - k_\varepsilon e^2_\eta e_\varepsilon^Te_\varepsilon -\sum_{i\in\mathcal{N}}k_{v_i}e_v^Te_v, \label{eq:V_dot}
\end{equation}
which is non-positive. We conclude therefore
that the system is stable and $V$ is a non-increasing function, deducing
the boundedness of $e_p,e_\eta,e_\varepsilon,e_v.$
Hence, invoking also the boundedness of $p_{\scriptscriptstyle O,\text{d}}$
and $\omega_{\scriptscriptstyle O,\text{d}}$ and of their
derivatives, we employ \eqref{eq:reference signals} to prove the
boundedness of $v_{\scriptscriptstyle O}^{\scriptscriptstyle r}$
and \eqref{eq:reference error} to prove the boundedness of $v_{\scriptscriptstyle O}$
and therefore of $v_{\scriptscriptstyle i},$ since
the boundedness of $J_{\scriptscriptstyle O_i}$
and $G$ is straightforward.
From the aforementioned conclusions, invoking also the fact that $M_i(\cdot),C_i(\cdot),M_{\scriptscriptstyle O}(\cdot), g_i(\cdot), C_{\scriptscriptstyle O}(\cdot), g_{\scriptscriptstyle O}(\cdot)$ are continuous functions, we can
deduce the boundedness of $q_i$ and $\dot{q}_{i},\forall i\in\mathcal{N}$ and of $\tilde{M},\tilde{C},\tilde{g}$.
Moreover, the error derivatives \eqref{eq:position error dynamics}-\eqref{eq:epsilon error dynamics}
are all bounded and thus, in view of \eqref{eq:reference signals}, $\dot{v}_{\scriptscriptstyle O}^{\scriptscriptstyle r}$
is bounded as well. Hence, we also deduce the boundedness
of $u_i$. Finally, by differentiating \eqref{eq:reference error}
and substituting \eqref{eq:coupled dynamics} and \eqref{eq:control law vector form}
we also deduce the boundedness of $\dot{e}_v$ and
therefore of $\dot{v}_{\scriptscriptstyle O}.$

Combining the aforementioned statements we can conclude the boundedness
of $\ddot{V}$ and hence the uniform continuity of $\dot{V}$. Invoking Barbalat's
lemma \citep{slotine1991applied}, we deduce that $\dot{V}\rightarrow0$ and therefore through
\eqref{eq:V_dot} that $(e_p,e_\eta e_\varepsilon,e_v)\rightarrow(0_{\scriptscriptstyle 3\times1},0_{\scriptscriptstyle 3\times1},0_{\scriptscriptstyle 6\times1})$. The equilibrium $e_\eta = 0$ can be proven to be unstable \citep{mayhew2011quaternion} and hence, since $e_\eta(0) \neq 0$, we conclude that $(e_p,e_\varepsilon,e_v)\rightarrow(0_{\scriptscriptstyle 3\times1},0_{\scriptscriptstyle 3\times1},0_{\scriptscriptstyle 6\times1})$.
Furthermore, we also conclude that $e_\eta^2\rightarrow1$ since
$e_\xi\in S^3$ is a unit quaternion, which leads to the completion of the proof. 
\end{pf}

\begin{remark}
The assumption $e_\eta(0)\neq 0$ is a necessary assumption to guarantee asymptotic stability of the orientation error $e_\xi$. In terms of Euler angles, it states that the initial orientation errors (in the $x,y,z$ directions) should not be $180$ degrees.
\end{remark}

\subsection{Adaptive Control Scheme}

Consider now that the dynamic parameters of the object
and the agents (e.g., masses and inertia moments), are unknown. We
propose an adaptive version of \eqref{eq:control law vector form}
that does not incorporate the aforementioned parameters and still
guarantees the solution of Problem \ref{prob:problem1}. 

It can be shown \citep{siciliano2010robotics} that the object and agent dynamics can be written in the form:
\begin{subequations} \label{eq: parameter linearity }
\begin{align}
& M_i(q_i)\dot{v}_i + C_i(q_i,\dot{q}_i)v_i + g_i(q_i) =  H_i(q_i,\dot{q}_i,v_i, \dot{v}_i)\theta_i \\
& M_{\scriptscriptstyle O}(x_{\scriptscriptstyle O})\dot{v}_{\scriptscriptstyle O} + C_{\scriptscriptstyle O}(x_{\scriptscriptstyle O},\dot{x}_{\scriptscriptstyle O})v_{\scriptscriptstyle O} + g_{\scriptscriptstyle O}(x_{\scriptscriptstyle O}) = \notag \\ 
& \hspace{4.4cm}  Y_{\scriptscriptstyle O}(x_{\scriptscriptstyle O},\dot{x}_{\scriptscriptstyle O},v_{\scriptscriptstyle O}, \dot{v}_{\scriptscriptstyle O})\theta_{\scriptscriptstyle O}, 
\end{align}
\end{subequations} 
$\forall i\in\mathcal{N}$, where $\theta_i \in\mathbb{R}^\ell, \theta_{\scriptscriptstyle O}\in\mathbb{R}^{\ell_{\scriptscriptstyle O}}$ are vectors of unknown but constant dynamic parameters of the agents and the object, appearing in the terms $M_i,C_i,g_i$ and $M_{\scriptscriptstyle O},C_{\scriptscriptstyle O},g_{\scriptscriptstyle O}$, respectively, and 
$H_i\in\mathbb{R}^{6\times\ell},i\in\mathcal{N}, Y_{\scriptscriptstyle O}\in\mathbb{R}^{6\times\ell_{\scriptscriptstyle O}}$ are known regressor matrices, independent of $\theta_i,\theta_{\scriptscriptstyle O}$. It is worth noting that
the choice for $\ell$ and $\ell_{\scriptscriptstyle O}$ is not unique and depends on the factorization method used \citep{siciliano2010robotics}. In the same vein, since $J_{\scriptscriptstyle O_i}$, as given in \eqref{eq:J_o_i_def}, depends only on $q_i$ and not on $\theta_i,\theta_{\scriptscriptstyle O},\forall i\in\mathcal{N}$, we can write:
\begin{align}
J^T_{\scriptscriptstyle O_i}M_iJ_{\scriptscriptstyle O_i}\dot{v}_i + (J^T_{\scriptscriptstyle O_i}M_i\dot{J}_{\scriptscriptstyle O_i} + J^T_{\scriptscriptstyle O_i}C_iJ_{\scriptscriptstyle O_i})v_i + J^T_{\scriptscriptstyle O_i}g_i = \notag \\ Y_i(q_i,\dot{q}_i,v_i,\dot{v}_i)\theta_i, \label{eq:tmp}
\end{align}
where $Y_i\in\mathbb{R}^{6\times\ell}$ is another regressor matrix independent of $\theta_i,\theta_{\scriptscriptstyle O}$. Hence, in view of \eqref{eq:coupled terms}, \eqref{eq: parameter linearity } and \eqref{eq:tmp}, the left-hand side of \eqref{eq:coupled dynamics} can be written as: 	
\begin{align}
\hspace{-2.2mm} \tilde{M}({q},x_{\scriptscriptstyle O})\dot{v}_{\scriptscriptstyle O}+ & \tilde{C}({q},\dot{{q}},x_{\scriptscriptstyle O},\dot{x}_{\scriptscriptstyle O})v_{\scriptscriptstyle O}+\tilde{g}({q},x_{\scriptscriptstyle O}) = \notag \\ & Y_{\scriptscriptstyle O}(x_{\scriptscriptstyle O},\dot{x}_{\scriptscriptstyle O},v_{\scriptscriptstyle O},\dot{v}_{\scriptscriptstyle O})\theta_{\scriptscriptstyle O} + Y^T({q},\dot{{q}},v_{\scriptscriptstyle O},\dot{v}_{\scriptscriptstyle O})\theta \label{eq:parameter linearity 2}
\end{align}
where $Y({q},\dot{{q}},v_{\scriptscriptstyle O}, \dot{v}_{\scriptscriptstyle O}) = [Y_1(q_1,\dot{q}_1,v_{\scriptscriptstyle O},\dot{v}_{\scriptscriptstyle O}),\dots,Y_N(q_N,\dot{q}_N,\\v_{\scriptscriptstyle O},\dot{v}_{\scriptscriptstyle O})]^T \in\mathbb{R}^{N\ell\times 6}$ and $\theta = [[\theta^T_i]_{i\in\mathcal{N}}]^T\in\mathbb{R}^{N\ell}$.

Let us now denote as $\hat{\theta}^i_{\scriptscriptstyle O}:\mathbb{R}_{\geq 0}\to\mathbb{R}^{\ell_{\scriptscriptstyle O}}$ and $\hat{\theta}_i:\mathbb{R}_{\geq 0}\to\mathbb{R}^{\ell}$ the estimates of $\theta_{\scriptscriptstyle O}$ and $\theta_i$, respectively, by agent $i\in\mathcal{N}$, and the corresponding stack vectors 
$\hat{\theta}_{\scriptscriptstyle O}(t) = [[(\hat{\theta}^i_{\scriptscriptstyle O}(t))^T]_{i\in\mathcal{N}}]^T\in\mathbb{R}^{N\ell_{\scriptscriptstyle O}}, \hat{\theta}(t) = [[\hat{\theta}^T_i(t)]_{i\in\mathcal{N}}]^T\in\mathbb{R}^{N\ell}$,
for which we formulate the associated errors $e_{\theta_{\scriptscriptstyle O}}:\mathbb{R}_{\geq 0}\to\mathbb{R}^{N\ell_{\scriptscriptstyle O}}, e_{\theta}:\mathbb{R}_{\geq 0}\to\mathbb{R}^{N\ell}$ as 
\begin{subequations} \label{eq:adaptation errors}
\begin{align}
e_{\theta_{\scriptscriptstyle O}}(t) = & \begin{bmatrix}
e^1_{\theta_{\scriptscriptstyle O}}(t) \\ \vdots \\ e^N_{\theta_{\scriptscriptstyle O}}(t) 
\end{bmatrix} = 
\begin{bmatrix}
\theta_{\scriptscriptstyle O} - \hat{\theta}^1_{\scriptscriptstyle O}(t) \\ \vdots \\ \theta_{\scriptscriptstyle O} - \hat{\theta}^N_{\scriptscriptstyle O}(t) 
\end{bmatrix}
 = \bar{\theta}_{\scriptscriptstyle O} - \hat{\theta}_{\scriptscriptstyle O}(t)   \label{eq:adaptation errors_1}\\
e_{\theta}(t) = & \begin{bmatrix}
e_{\theta_1}(t) \\ \vdots \\ e_{\theta_N}(t)
\end{bmatrix}  = 
\begin{bmatrix}
\theta_1- \hat{\theta}_1(t) \\ \vdots \\ \theta_N- \hat{\theta}_N(t)
\end{bmatrix} = \theta- \hat{\theta}(t),   \label{eq:adaptation errors_2}
\end{align}
\end{subequations}
$\forall i\in\mathcal{N}$, where $\bar{\theta}_{\scriptscriptstyle O}= [\underbrace{\theta^T_{\scriptscriptstyle O}, \dots, \theta^T_{\scriptscriptstyle O}}_{N \text{times}}]^T\in\mathbb{R}^{N\ell_{\scriptscriptstyle O}}$.

Then, with the reference velocity signal $v^{\scriptscriptstyle r}_{\scriptscriptstyle O}$ defined as in \eqref{eq:reference signals}
and the corresponding error $e_v$  as in \eqref{eq:reference error}, we
design the adaptive control law $u_i:\mathbb{R}_{\geq 0}\to\mathbb{R}^6$ in \eqref{eq:coupled dynamics}, for each agent $i\in\mathcal{N}$, as:
\begin{align}
u_i(t) =& J^{-T}_{\scriptscriptstyle O_i}\left(Y_i(q_i,\dot{q}_i,v^{\scriptscriptstyle r}_{\scriptscriptstyle O},\dot{v}^{\scriptscriptstyle r}_{\scriptscriptstyle O})\hat{\theta}_i(t)  - c_ie(t) - k_{v_i}e_v(t) \notag \right. \\
& \left. + c_iY_{\scriptscriptstyle O}(x_{\scriptscriptstyle O},\dot{x}_{\scriptscriptstyle O},v^{\scriptscriptstyle r}_{\scriptscriptstyle O},\dot{v}^{\scriptscriptstyle r}_{\scriptscriptstyle O})\hat{\theta}^i_{\scriptscriptstyle O}(t)  \right),	\notag
\end{align}
which can be written in vector form as
\begin{align}
{u}(t) = \tilde{G}^T\left(\tilde{Y}(\cdot)\hat{\theta}(t) + \tilde{Y}_{\scriptscriptstyle O}(\cdot)\hat{\theta}_{\scriptscriptstyle O}(t) - C_fe(t) - {K}_ve_v(t) \right), \label{eq:adaptive control vector form}
\end{align}
where $\tilde{Y}(\cdot) = \text{diag}\{[Y_i(q_i,\dot{q}_i,v^{\scriptscriptstyle r}_{\scriptscriptstyle O},\dot{v}^{\scriptscriptstyle r}_{\scriptscriptstyle O})]_{i\in\mathcal{N}}  \}\in\mathbb{R}^{6N\times N\ell}$, $\tilde{Y}_{\scriptscriptstyle O}(\cdot) = \text{diag}\{[c_iY_{\scriptscriptstyle O}(x_{\scriptscriptstyle O},\dot{x}_{\scriptscriptstyle O},v^{\scriptscriptstyle r}_{\scriptscriptstyle O},\dot{v}^{\scriptscriptstyle r}_{\scriptscriptstyle O})]_{i\in\mathcal{N}}\}\in\mathbb{R}^{6N\times N\ell_{\scriptscriptstyle O}}, \tilde{G}$, $C_f$, ${K}_v$ as defined in \eqref{eq:control law vector form}, and 
$e$ as defined in \eqref{eq:reference signals}.

Moreover, we design the adaptation laws $\dot{\hat{\theta}}^i_{\scriptscriptstyle O}:\mathbb{R}_{\geq 0}\to\mathbb{R}^{\ell_{\scriptscriptstyle O}}, \dot{\hat{\theta}}_i:\mathbb{R}_{\geq 0}\to\mathbb{R}^{\ell}, i\in\mathcal{N}$, for each agent, as: 
\begin{align}
\dot{\hat{\theta}}^i_{\scriptscriptstyle O}(t) =& -c_i Y^T_{\scriptscriptstyle O}(x_{\scriptscriptstyle O},\dot{x}_{\scriptscriptstyle O},v^{\scriptscriptstyle r}_{\scriptscriptstyle O},\dot{v}^{\scriptscriptstyle r}_{\scriptscriptstyle O})e_v(t) \notag\\
\dot{\hat{\theta}}_i(t) = & -\gamma_iY^T_i(q_i,\dot{q}_i,v^{\scriptscriptstyle r}_{\scriptscriptstyle O},\dot{v}^{\scriptscriptstyle r}_{\scriptscriptstyle O})e_v(t), \notag
\end{align}  
which is written in vector form as 
\begin{subequations} \label{eq:adaptation laws}
\begin{align}
\dot{\hat{\theta}}_{\scriptscriptstyle O}(t) = & \begin{bmatrix}
\dot{\hat{\theta}}^1_{\scriptscriptstyle O}(t) \\ \vdots \\ \dot{\hat{\theta}}^N_{\scriptscriptstyle O}(t)
\end{bmatrix} = 
-\tilde{C}_fY^T_{\scriptscriptstyle O}(x_{\scriptscriptstyle O},\dot{x}_{\scriptscriptstyle O},v^{\scriptscriptstyle r}_{\scriptscriptstyle O},\dot{v}^{\scriptscriptstyle r}_{\scriptscriptstyle O})e_v(t) \\ 
\dot{\hat{\theta}}(t) = & \begin{bmatrix}
\dot{\hat{\theta}}_1(t) \\ \vdots \\ \dot{\hat{\theta}}_N(t)
\end{bmatrix} = -\Gamma Y({q},\dot{{q}},x_{\scriptscriptstyle O},\dot{x}_{\scriptscriptstyle O})e_v(t),  	
\end{align}
\end{subequations}
where $\Gamma = \text{diag}\{[\gamma_i I_{\ell}]_{i\in\mathcal{N}}\}\in\mathbb{R}^{N\ell\times N\ell}_{\geq 0}, \gamma_i\in\mathbb{R}_{>0}$, and $\tilde{C}_f = [c_1 I_{\ell_{\scriptscriptstyle O}}, \dots, c_N I_{\ell_{\scriptscriptstyle O}}]^T\in\mathbb{R}^{N\ell_{\scriptscriptstyle O}\times\ell_{\scriptscriptstyle O}}$.

The following theorem summarizes the main results of this subsection.

\begin{thm}
Consider $N$ robotic agents rigidly grasping an object with coupled
dynamics described by (\ref{eq:coupled dynamics}) and unknown dynamic
parameters. Then, by applying the control protocol \eqref{eq:adaptive control vector form}
with the adaptation laws \eqref{eq:adaptation laws}, and under the assumption $e_\eta(0)\neq 0$,
the object pose converges asymptotically to the desired pose with
all closed loop signals being bounded, i.e, Problem \ref{prob:problem1}
is solved. 
\end{thm}
\begin{pf}
Consider the positive definite, decrescent and radially unbounded
Lyapunov function
\begin{align}
V(e_p,e_\eta, e_v,e_\theta, e_{\theta_{\scriptscriptstyle O}},t) =  \tfrac{1}{2}e_p^Te_p+e^2_\eta-1+\notag\\  \tfrac{1}{2}e^T_v\tilde{M}(q(t),x_{\scriptscriptstyle O}(t))e_v+\tfrac{1}{2}e^T_\theta \Gamma^{-1} e_\theta + \tfrac{1}{2}e^T_{\theta_{\scriptscriptstyle O}}e_{\theta_{\scriptscriptstyle O}}, \notag
\end{align}
which, by time differentiation and by employing 
the error dynamics \eqref{eq:error_dynamics}, relations 
\eqref{eq:reference signals},
\eqref{eq:reference error} as well as \eqref{eq:parameter linearity 2} and \eqref{eq:adaptation errors},
becomes: 
\begin{align}
\dot{V} = & -e^{T}Ke+e^T_v(e - Y_{\scriptscriptstyle O}(x_{\scriptscriptstyle O}, \dot{x}_{\scriptscriptstyle O},v^{\scriptscriptstyle r}_{\scriptscriptstyle O},\dot{v}^{\scriptscriptstyle r}_{\scriptscriptstyle O})\theta_{\scriptscriptstyle O} - \notag\\ 
& Y^T({q},\dot{{q}},v^{\scriptscriptstyle r}_{\scriptscriptstyle O},\dot{v}^{\scriptscriptstyle r}_{\scriptscriptstyle O})\theta + G^T{u}) - e^T_{\theta_{\scriptscriptstyle O}}\dot{\hat{\theta}}_{\scriptscriptstyle O} - e^T_\theta\Gamma^{-1}\dot{\hat{\theta}}. \notag
\end{align}
By substituting the adaptive control protocol \eqref{eq:adaptive control vector form}, and noticing that, due to the fact that $\sum_{i\in\mathcal{N}}c_i = 1$, it holds that $Y_{\scriptscriptstyle O}(\cdot)\theta_{\scriptscriptstyle O} = Y_{\scriptscriptstyle O}(\cdot)\tilde{C}^T_f\bar{\theta}_{\scriptscriptstyle O}$, we obtain: 
\begin{align}
\dot{V} = & -e^TKe-\sum_{i\in\mathcal{N}}k_{v_i}e_v^Te_v - e^T_vY_{\scriptscriptstyle O}(\cdot)\tilde{C}^T_fe_{\theta_{\scriptscriptstyle O}} - e_vY^T(\cdot)e_\theta \notag \\
& - e^T_{\theta_{\scriptscriptstyle O}}\dot{\hat{\theta}}_{\scriptscriptstyle O} - e^T_\theta\Gamma^{-1}e_\theta, \notag	
\end{align}
which, after substituting the adaptive laws \eqref{eq:adaptation laws}, becomes
\begin{equation}
\dot{V}= -k_pe_p^Te_p - k_\varepsilon e_\eta e_\varepsilon^Te_\varepsilon -\sum_{i\in\mathcal{N}}k_{v_i}e_v^Te_v, \notag
\end{equation}
which is non-positive. We conclude therefore
the boundedness of $V$ and of $e_j,j\in\{ p,\eta,\varepsilon,v,\theta_{\scriptscriptstyle O},\theta\}$ 
and hence the boundedness of $v_{\scriptscriptstyle O}^{\scriptscriptstyle r},v_{\scriptscriptstyle O},v_i,\hat{\theta}_{\scriptscriptstyle O}$
and $\hat{\theta}$. By proceeding in a similar manner
as in the non-adaptive scenario, we prove the boundedness of ${\dot{v}_{{\scriptscriptstyle O}}^{{\scriptscriptstyle r}},\dot{v}_{{\scriptscriptstyle O}}}$
and of ${\dot{e}_{p},\dot{e}_{\varepsilon},\dot{e}_{v}}.$
We deduce, therefore, the boundedness of $\ddot{V}$ and consequently the
uniform continuity of $\dot{V}$ and that $(e_p,e_\varepsilon,e_v)\rightarrow(0_{3\times1},0_{3\times1},0_{6\times1})$ and hence, $e^2_\eta \to 1$.
Finally, it can be proved that the control and adaptation
signals ${{u}},{\dot{\hat{\theta}}_{{\scriptscriptstyle O}}},{\dot{\hat{\theta}}}$
are also bounded, which leads to the completion of the proof. \end{pf}

\begin{remark}
Note that the dynamic parameter errors $e_{\theta_{\scriptscriptstyle O}}, e_{\theta}$ are only guaranteed to stay bounded, not to be asymptotically driven to zero. However, that does not affect the result of the aforementioned analysis that $(e_p, e_\varepsilon, \lvert e_\eta \rvert, e_v) \to (0_{3\times1},0_{3\times1},1, 0_{6\times1})$. 
Moreover, the gains $k_p, k_\varepsilon$ in \eqref{eq:reference signals} must be known by all agents $i\in\mathcal{N}$, and the  
load-sharing coefficients $c_i,i\in\mathcal{N}$ cannot be arbitrarily chosen by each agent, due to the constraint that $\sum_{i\in\mathcal{N}}c_i = 1$. Nevertheless, these values are constant and can be transmitted off-line to the agents.
\end{remark}

\begin{remark}
In both control methodologies \eqref{eq:control law vector form},\eqref{eq:adaptive control vector form}, we can guarantee internal force regulation by including a vector of desired internal forces ${f}_{\text{int,d}}:\mathbb{R}_{\geq 0}\to\mathbb{R}^{6N}$ that belong to the nullspace of $G^T$, i.e., ${f}_{\text{int,d}}(t) = (I_{6N} - G^*G^T)\hat{{f}}_{\text{int,d}}$, where $G^*:\mathbb{R}^n\to\mathbb{R}^{6N\times6}$, with $G^*({q}) = \tfrac{1}{N}[J^{-1}_{\scriptscriptstyle O_1}(q_1),\dots,J^{-1}_{\scriptscriptstyle O_N}(q_N)]^T$, and $\hat{{f}}_{\text{int,d}}$ a constant vector that can be transmitted off-line to the agents. It can be proved then, in view of the aforementioned analysis, that when $t\to\infty$, the generalized force vector acting on the object's center of mass consists of a term that results in its motion (for the trajectory tracking), and the term associated with the internal forces.
Note though, that the computation of $G^*G^T$ requires knowledge of all grasping points $p_{\scriptscriptstyle E_i}$, which reduces to knowledge of the constant offsets $p^{\scriptscriptstyle O }_{\scriptscriptstyle E_i/O}$, since, from \eqref{eq:coupled_kinematics_1}, we have that $p_{\scriptscriptstyle E_i}(t) = p_{\scriptscriptstyle O}(t) + R_{\scriptscriptstyle O}p^{\scriptscriptstyle O}_{\scriptscriptstyle E_i/O}$ and therefore, each agent can compute all $p_{\scriptscriptstyle E_i}(t),\forall i\in\mathcal{N}$, since it can compute the pose of the object's center of mass. Therefore, by off-line transmission of all $p^{\scriptscriptstyle O}_{\scriptscriptstyle E_i/O}$ to all agents, internal force regulation can be achieved. 
\end{remark} 

\begin{figure}
\begin{center}
\includegraphics[width = 0.47\textwidth]{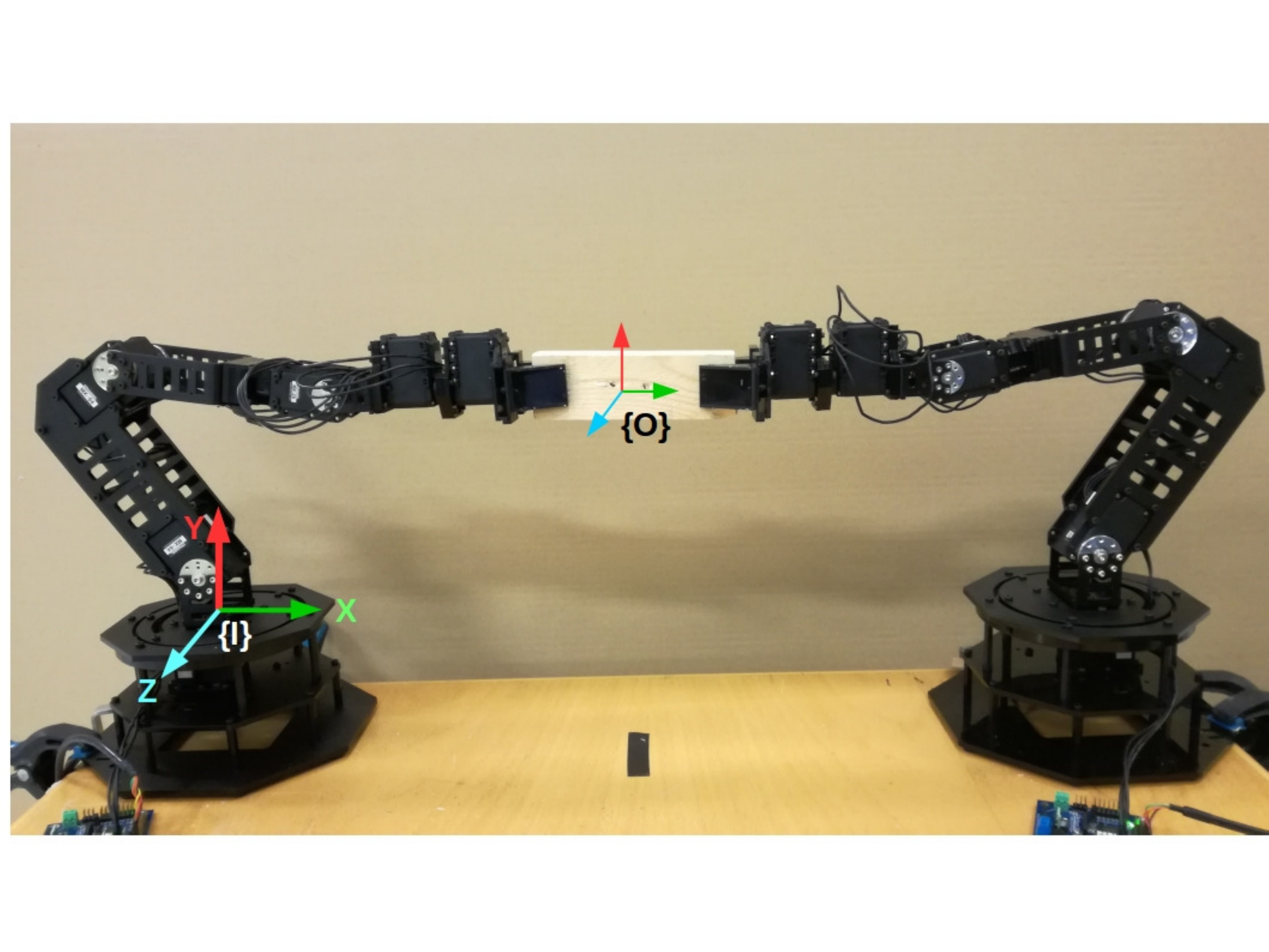}
\caption{Two WidowX Robot Arms rigidly grasping an object; $\{I\}$ and $\{O\}$ denote the inertial and the object's frame, respectively. \label{fig:exp_arms}}
\end{center}
\end{figure}

\section{Experimental Evaluation} \label{sec:Experiments}
To demonstrate the efficiency of the proposed algorithm, en experimental study was carried out using two WidowX Robot Arms, as shown in Fig. \ref{fig:exp_arms}, and the non-adaptive version of the proposed control scheme. The desired profile to be tracked by the object was determined by the planar motion $p_{\scriptscriptstyle O,\text{d}}(t) = [0.3+0.05\sin(\tfrac{2\pi}{15}t),0.12,0]^T\text{m}$ and $\xi_{\scriptscriptstyle O,\text{d}}(t) = [\eta_{\scriptscriptstyle O,\text{d}}(t),\varepsilon^T_{\scriptscriptstyle O,\text{d}}(t)]^T =  [\cos(\tfrac{\pi}{60}\sin(\tfrac{2\pi}{15}t)),0,0,-\sin(\tfrac{\pi}{60}\sin(\tfrac{2\pi}{15}t))]^T$, that is associated to the angle trajectory   $\phi_{\scriptscriptstyle O,\text{d}}(t) = [0,0,-\tfrac{\pi}{30}\sin(\tfrac{2\pi}{15}t)]^T \text{rad}$ with respect to the $z$ axis. For the execution of the task, we employed the three rotational joints with respect to the $z$-axis (see Fig. \ref{fig:exp_arms}) of the arms. 
The object's initial pose was $p_{\scriptscriptstyle O}(0) = [0.301,0.123,0]^T\text{m}, \phi_{\scriptscriptstyle O}(0) = [0,0,0]^T\text{rad}$. The load sharing coefficients and the control gains were chosen as $c_1 = c_2 = 0.5$  and $k_p = 150, k_\varepsilon = 100, k_{v_1} = k_{v_2} = 2.5$, respectively. 

\begin{figure}
\begin{center}
\includegraphics[width = 0.47\textwidth]{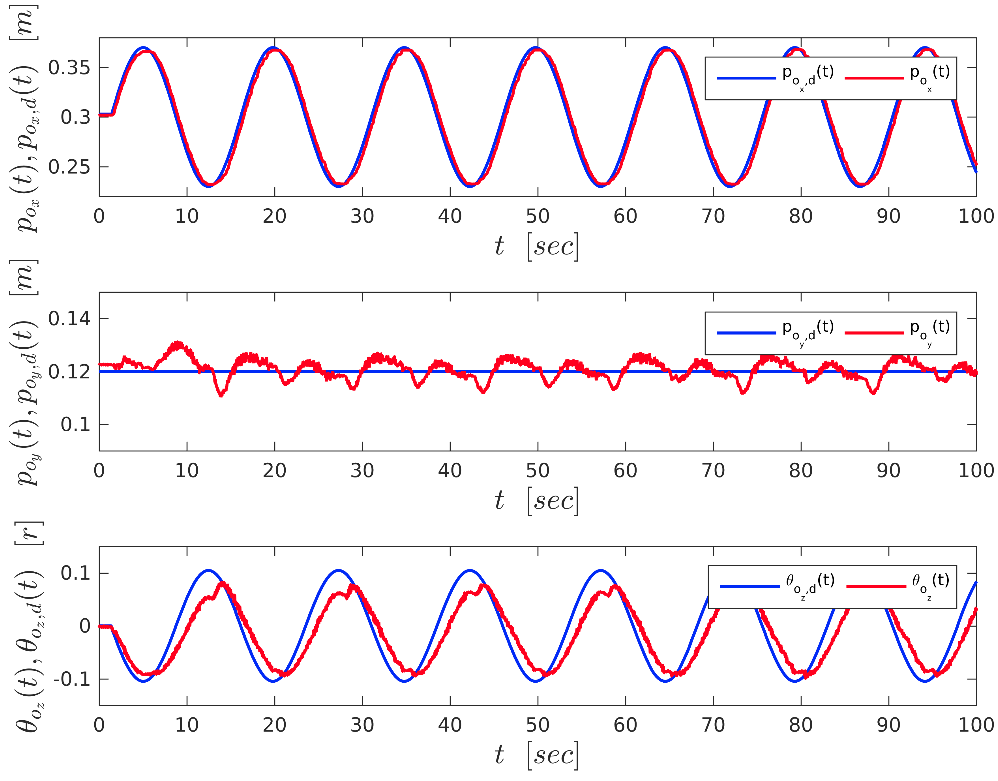}
\caption{The desired (with blue) and actual (with red) pose trajectory of the object's center of mass $p_{\scriptscriptstyle O,\text{d}}(t)$ and $p_{\scriptscriptstyle O}(t)$, respectively, for $t\in[0,100]\text{s}$. Top: $x$ (horizontal) direction. Middle: $y$ (vertical) direction. Bottom: Angle $\phi_{\scriptscriptstyle O_z}(t)$ with respect to $z$ axis (direction perpendicular to plane $x$-$y$).  \label{fig:p_p_des}}
\end{center}
\end{figure}

\begin{figure}
\begin{center}
\includegraphics[width = 0.47\textwidth]{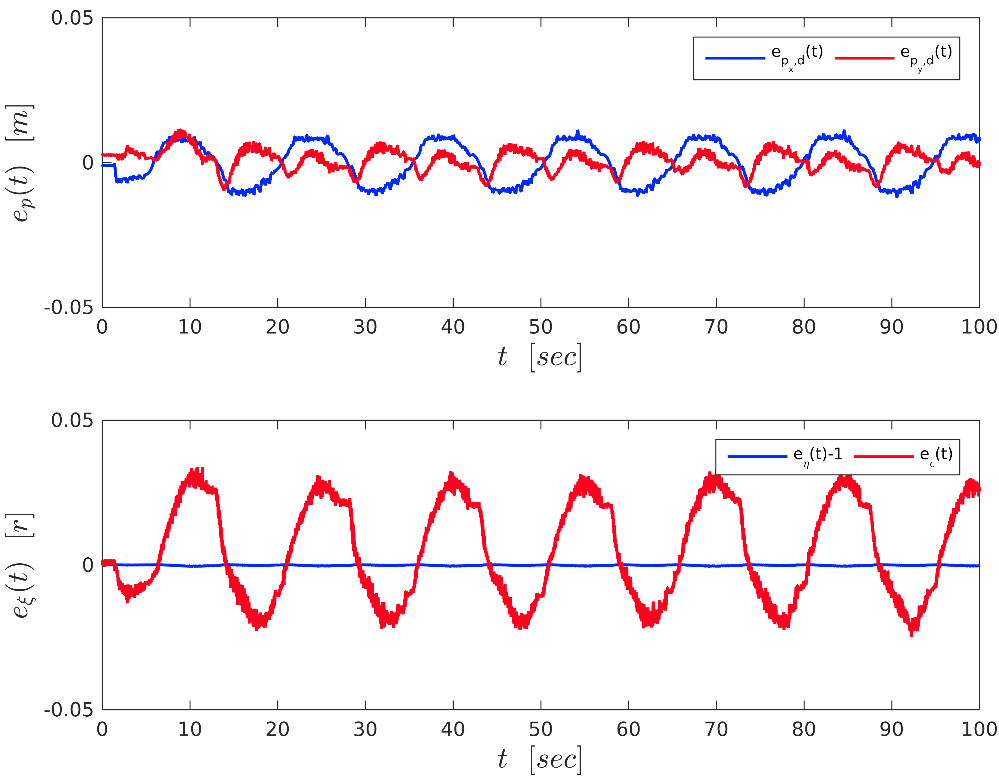}
\caption{The evolution of the errors $e_p(t), e_{\xi}(t), t\in[0,100]\text{s}$. Top: $e_p(t)$ in $x$ (with blue) and $y$ (with red) direction. Bottom: $e_\eta(t)-1$ (with blue) and $e_\varepsilon(t)$ (with red).   \label{fig:errors}}
\end{center}
\end{figure}

\begin{figure}
\begin{center}
\includegraphics[width = 0.47\textwidth]{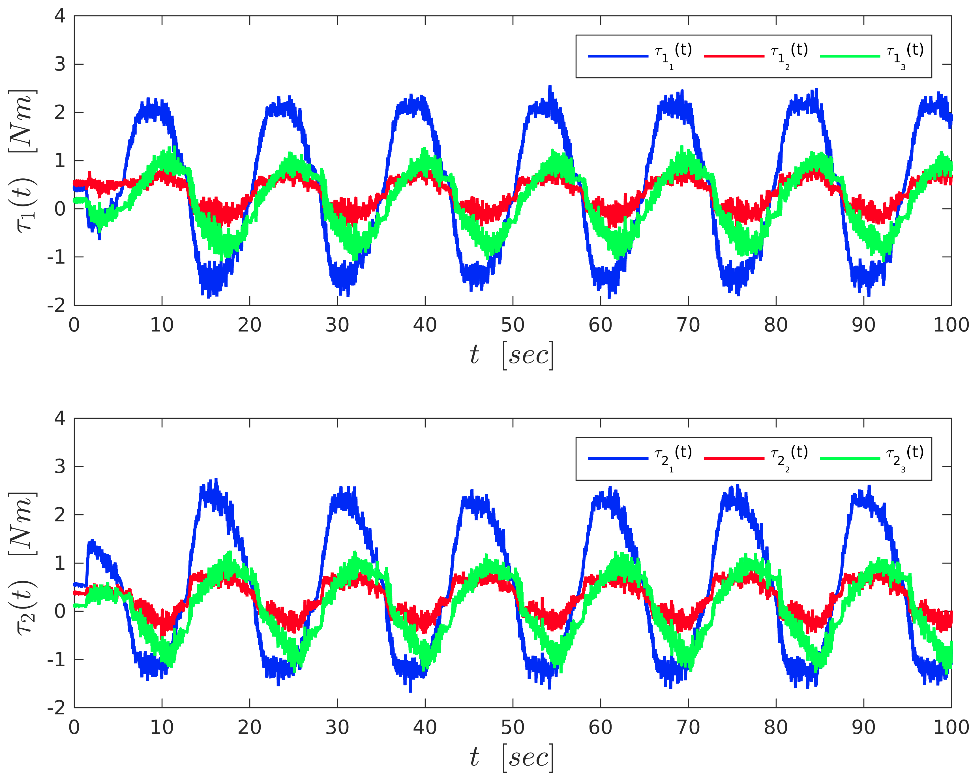}
\caption{The resulting input torques $\tau_1(t)$ (top) and $\tau_2(t)$ (bottom), $t\in[0,100]\text{s}$. \label{fig:tau}}
\end{center}
\end{figure}

\begin{figure}
\begin{center}
\includegraphics[width = 0.47\textwidth]{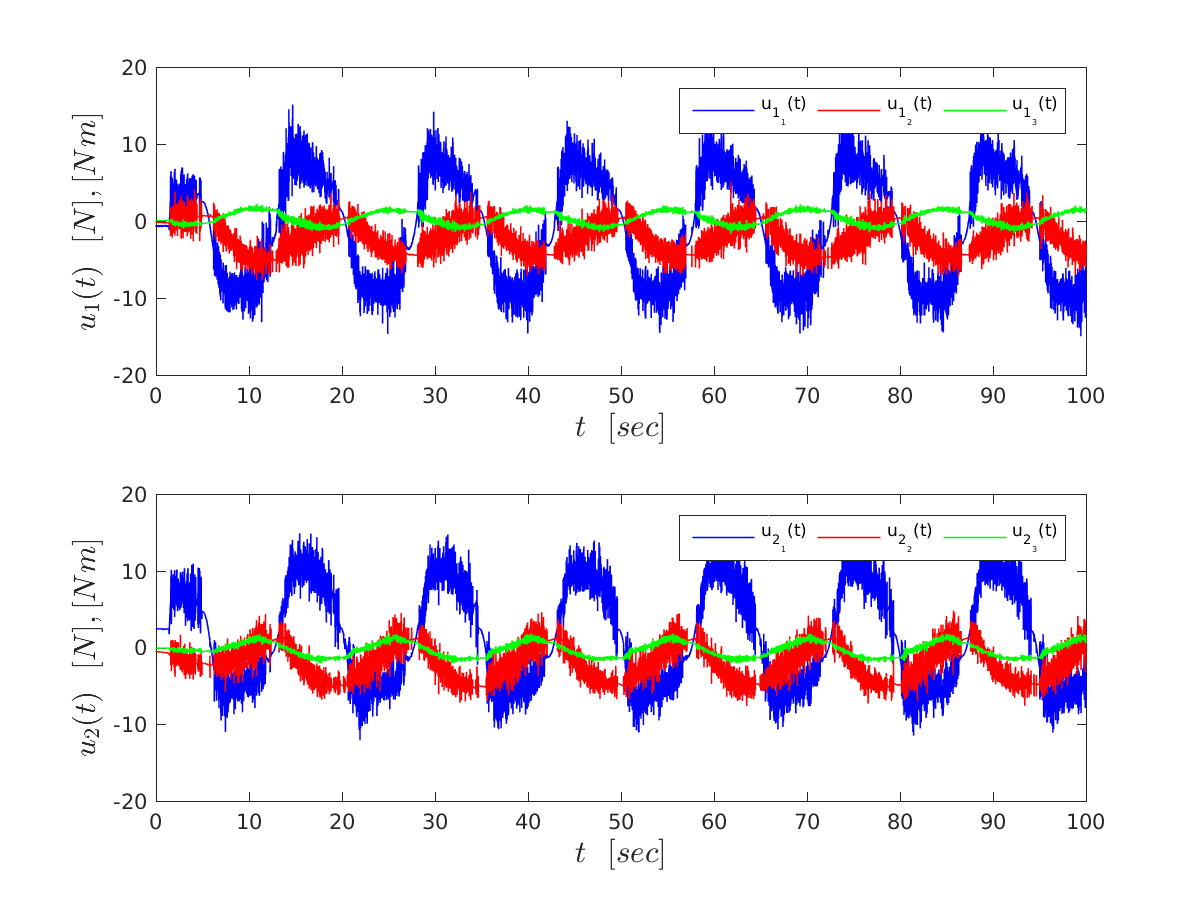}
\caption{The resulting $3$D task-space wrenches $u_1(t)$ (top) and $u_2(t)$ (bottom), $t\in[0,100]\text{s}$. \label{fig:f}}
\end{center}
\end{figure}

The experimental results for $t=10^2$s are depicted in Fig. \ref{fig:p_p_des}-\ref{fig:f}. In particular, the tracking of the desired object pose by the actual one is illustrated in Fig. \ref{fig:p_p_des}. Moreover, the evolution of the errors $e_p(t)$ and $e_\xi(t)$ is depicted in Fig. \ref{fig:errors}. It can be concluded from the figures that the tracking of the desired pose is achieved with some negligible oscillatory behavior that can be attributed to the deviation of the dynamics \eqref{eq:coupled dynamics} from the actual coupled dynamics due to sensor noise, unmodelled friction, external disturbances and small sliding in the contact points which affects the rigidity assumption. The torque signals $\tau_i,i\in\{1,2\}$ are pictured in Fig \ref{fig:tau} and the task-space wrenches $u_i,i\in\{1,2\}$ in Fig. \ref{fig:f}. It can be seen that the proposed algorithm does not output large values for the resulting input torques and forces. A short video demonstrating the experimental setup can be found at https://youtu.be/PCnZ6C8ECFg.

\section{Conclusion} \label{sec:Conclusion}

We have proposed a novel control protocol for the cooperative manipulation of an object by $N$ robotic agents using unit quaternions and without employing any force/torque measurements. Future efforts will be devoted towards incorporating kinematic uncertainties associated with the location of the object's center of mass, external disturbances, non-rigid grasps as well as singularity avoidance.

\bibliography{ifacconf}             
                                                   







\end{document}